\documentclass[pdflatex,sn-nature]{sn-jnl}
\usepackage[utf8]{inputenc}
\theoremstyle{thmstyleone}%
\theoremstyle{thmstyletwo}%

\theoremstyle{thmstylethree}%

\usepackage{graphicx}%
\usepackage{multirow}%
\usepackage{amsmath,amssymb,amsfonts}%
\usepackage{amsthm}%
\usepackage{mathrsfs}%
\usepackage[title]{appendix}%
\usepackage{xcolor}%
\usepackage{textcomp}%
\usepackage{manyfoot}%
\usepackage{booktabs}%
\usepackage{algorithm}%
\usepackage{algorithmicx}%
\usepackage{algpseudocode}%
\usepackage{listings}%
\usepackage{graphicx}
\graphicspath{{figures/}} 

\begin{document}

\title[Short Title]{Kinematic and Ergonomic Design of a Robotic Arm for Precision Laparoscopic Surgery}

\author*[1]{\fnm{Tian} \sur{Hao}}\email{htian@ust.hk}
\author[2]{\fnm{Tong} \sur{Lu}}\email{lut6@mail.uc.edu}
\author[1]{\fnm{Che} \sur{Chan}}\email{cchan@ust.hk}

\affil*[1]{\orgdiv{Department of Computer Science}, \orgname{Hongkong University of Science and Technology}}
\affil[2]{\orgdiv{Department of Computer Science}, \orgname{UC}}


\maketitle

\begin{abstract}
AAbstract: Robotic assistance in minimally invasive surgery can greatly enhance surgical precision and reduce surgeon fatigue. This paper presents a focused investigation on the kinematic and ergonomic design principles for a laparoscopic surgical robotic arm aimed at high-precision tasks. We propose a 7-degree-of-freedom (7-DOF) robotic arm system that incorporates a remote center of motion (RCM) at the instrument insertion point and ergonomic considerations to improve surgeon interaction. The design is implemented on a general-purpose robotic platform, and a series of simulated surgical tasks were performed to evaluate targeting accuracy, task efficiency, and surgeon comfort compared to conventional manual laparoscopy. Experimental results demonstrate that the optimized robotic design achieves significantly improved targeting accuracy (error reduced by over 50\%) and shorter task completion times, while substantially lowering operator muscle strain and discomfort. These findings validate the importance of kinematic optimization (such as added articulations and tremor filtering) and human-centered ergonomic design in enhancing the performance of robot-assisted surgery. The insights from this work can guide the development of next-generation surgical robots that improve surgical outcomes and ergonomics for the operating team.
\end{abstract}

\section{Introduction}\label{sec1}

Robot-assisted minimally invasive surgery has transformed surgical practice by enabling greater precision and dexterity through small incisions \cite{fuerst2015first}. Advanced imaging and guidance technologies are being integrated to further improve surgical accuracy \cite{duan2025localization}. For example, augmented reality (AR) visualization has recently been applied to intraoperative guidance, significantly reducing localization errors in tasks like sentinel lymph node biopsy \cite{von2024augmented}. Such high-precision requirements in modern surgery motivate a closer examination of the underlying robotic design principles that enable accurate and ergonomic instrument control.

A key aspect of surgical robot design is the kinematic configuration of the robotic arms that manipulate instruments inside the patient’s body \cite{kuo2012kinematic, ng2014current}. In laparoscopic surgery, instruments must pivot about a fixed entry point on the patient’s body wall---a constraint known as the \textit{remote center of motion} (RCM). Robotic systems like the da Vinci\textsuperscript{\textregistered} employ specialized linkages or control algorithms to maintain an RCM, allowing instruments to move pivotally without excessive force on the incision \cite{zhang2024state, colan2023manipulability}. Providing sufficient degrees of freedom (DOF) is crucial: a conventional laparoscopic tool effectively has 4 DOF (pitch, yaw, insertion, rotation) about the trocar point, whereas a robotic instrument with a wrist can restore the full 6-7 DOF of motion inside the body. The additional articulations can greatly enhance dexterity, enabling complex maneuvers such as suturing and knot-tying to be performed more easily and accurately \cite{li2022robotic}. However, adding DOF and maintaining rigidity pose engineering challenges in the arm’s design, requiring careful optimization of link lengths, joint placement, and actuation to achieve both precision and sufficient workspace \cite{kim2021control}.

Another critical design consideration is the \textit{ergonomics} of the surgical system \cite{dixon2024robotic, enayati2016haptics, moglia2021systematic}. Traditional manual laparoscopy forces surgeons to adopt uncomfortable postures and hand motions, often leading to fatigue or musculoskeletal strain during long procedures \cite{wong2024manipulation, wright2017robotic}. Robotic teleoperation consoles aim to improve the surgeon’s ergonomics by providing an intuitively controlled interface (e.g., joystick or master manipulators) and by filtering hand tremor and scaling motions \cite{perez2024comparative, gerhardus2003robot}. Nonetheless, the mechanical design of the patient-side robot also impacts the assistant’s and surgeon’s workflow \cite{cooper2025systematic, zhang2021ergonomic,qian2019review, talamini2003prospective}. Large, bulky robotic arms can clutter the operating room and create awkward angles for instrument insertion, whereas a more compact, human-centric design can facilitate smoother setup and lessen the physical burden on the surgical team. Thus, optimizing the robot’s form factor and alignment with the human anatomy is essential for both safety and ease of use.

In this work, we narrow our focus to the kinematic and ergonomic design principles of a robotic arm intended for precision laparoscopic surgery. The specific research question addressed is: \textit{How can a surgical robotic arm be designed to maximize targeting precision inside the patient while minimizing operator effort and discomfort?} We approach this problem by designing a 7-DOF robotic arm that integrates an RCM constraint and enhanced dexterity for instrument control, and by incorporating ergonomic features such as improved arm geometry and control positioning. The design is implemented on a general-purpose robotic platform (provided by Google’s robotics division) to evaluate its performance in comparison to a conventional laparoscopic technique. We hypothesize that the optimized robotic design will achieve significantly lower targeting error and faster task completion than manual operation, while also reducing the physical strain on the user.

To test this hypothesis, we conducted experiments in a simulated surgical setup, where participants performed fine targeting tasks using both the robotic system and a standard laparoscopic tool. Quantitative metrics of accuracy, efficiency, and user ergonomics were measured and analyzed. This paper presents the design of the robotic system (Section~\ref{sec2}), the experimental methods (Section~\ref{sec2}), the results of the comparative evaluation (Section~\ref{sec4}), and a discussion of the findings in the context of surgical robotics design (Section~\ref{sec5}). Our study provides evidence-based guidance for engineering surgical robots that can improve surgical precision while safeguarding the health and performance of the surgical staff.

\section{Materials and Methods}\label{sec2}

\subsection{Robotic Arm Design and Kinematics}\label{subsec2-1}

The robotic system in this study consists of a 7-DOF serial manipulator configured for laparoscopic surgery. The hardware platform is a general-purpose robotic arm developed by Google (Mountain View, CA, USA) with modifications for surgical instrument manipulation. The arm has a typical anthropomorphic arrangement (shoulder, elbow, wrist joints), providing a large workspace and human-like dexterity. We integrated a custom instrument holder at the end-effector that allows standard laparoscopic instruments (5~mm diameter tools) to be attached and controlled. A key feature of our design is the enforcement of a \textbf{Remote Center of Motion (RCM)} at the instrument insertion point. This ensures that the instrument shaft pivots about a fixed point on the patient's abdominal wall, mimicking the natural constraint of a trocar.

We achieve the RCM constraint via a software-controlled approach: the robot’s controller continuously solves inverse kinematics with the constraint that a particular point on the instrument (the fulcrum) remains stationary in space. This method, sometimes called “virtual RCM,” obviates the need for a dedicated mechanical linkage at the trocar, increasing flexibility. The 7-DOF arm provides redundancy, meaning there is an extra degree of freedom beyond the minimum required for the task. This redundancy was exploited to optimize arm posture; for example, the arm can adjust its elbow configuration without moving the instrument tip, allowing avoidance of joint limits or collisions with the operating table. The instrument itself adds an additional 1 DOF (rotation about its long axis), and the instrument tip features a wrist with 2 DOF (pitch and yaw) in our design. In total, the instrument tip can be manipulated with effectively 6 DOF of translation and orientation inside the body, equivalent to the dexterity of an open surgical tool, plus a gripper action for tissue manipulation.

We paid special attention to the precision of the arm’s kinematics. High-resolution encoders at each joint (with angular resolution $<0.01^\circ$) provide accurate feedback for fine motion control. The arm’s forward kinematics error was calibrated using a pivot calibration method: the robot moved the instrument in a small circle around the insertion point to ensure minimal lateral motion at the trocar. Any residual offset was corrected through software until the instrument could pitch and yaw around the RCM with errors $<1$~mm. This calibration is crucial because even minor deviations at the fulcrum can cause tissue stress or imprecise instrument positioning. Additionally, the structural stiffness of the arm was optimized through finite element analysis during design to minimize deflection under load. We targeted an end-effector deflection of less than 1~mm under typical forces (5~N) experienced during tissue manipulation. This stiffness ensures that the accuracy of the manipulator is not compromised by flexing when it applies force on the surgical site.

\subsection{Ergonomic Considerations in Design}\label{subsec2-2}

Beyond kinematics and accuracy, the design emphasizes ergonomic integration into the surgical workflow. One aspect of ergonomics is the physical \textbf{workspace and footprint} of the robot. Our robotic arm is relatively compact, with a reach of approximately 700~mm and a weight of 12~kg, allowing it to be mounted on a mobile cart or directly on the operating table railing. The base of the arm was designed to have a small footprint so it can be positioned close to the operating table without obstructing personnel. The arm can be easily draped for sterility and has a quick-release instrument mount so that tools can be exchanged in seconds, reducing the physical effort and time when switching instruments.

The control interface for the surgeon was a standard master console (similar to that of the da Vinci system), where the surgeon’s hand movements are translated to the instrument motions. We configured the scaling such that a 5:1 motion downscaling was applied (5 cm hand movement = 1 cm instrument movement), to allow very fine manipulation. Tremor filtering was implemented by averaging hand inputs over a 50~ms window, smoothing involuntary high-frequency motions. These settings were chosen based on known ergonomic benefits of robotic systems---for instance, motion scaling and tremor reduction help improve precision and reduce the cognitive and physical load on the surgeon’s hands
ales.amegroups.org
. The surgeon is seated comfortably at the console, which eliminates the need to stand in awkward positions as in traditional laparoscopy. The console includes an armrest and a stereoscopic display, further enhancing comfort and reducing fatigue.

For the assisting staff at the patient bedside, the robot’s slim design and articulating arm allow better access to the patient. We arranged the arm to approach the patient from an over-the-shoulder angle, which leaves the abdomen largely unobstructed for the assistant to perform tasks like suction or retraction if needed. The instrument insertion angle can be adjusted by the robot to be optimal for the surgical target area, which also means the entry port can be positioned to minimize musculoskeletal strain on the assistant (who might otherwise need to hold laparoscopic instruments in uncomfortable orientations). In this way, the robotic system not only benefits the surgeon at the console but also improves conditions for the entire team.

Finally, we incorporated safety and ergonomic feedback in the design. The system monitors forces at the instrument and provides haptic or visual feedback cues if excessive force is detected, prompting the user to adjust. This prevents the surgeon from unknowingly exerting high forces that could both harm the patient and cause stress through the instrument back to the robot and onto the patient’s body wall. An automatically adjusting clutch was included so that if the surgeon releases the master controls, the robot holds the instrument in place, avoiding any drift that could cause tissue damage. This feature allows the surgeon to take short breaks (micro-pauses) during lengthy tasks, a practice known to reduce fatigue
ales.amegroups.org
.

\subsection{Experimental Task and Protocol}\label{subsec2-3}

To evaluate the performance of our robotic arm design, we devised a simulated surgical task focusing on precision and comparing it to a manual laparoscopic approach. The task involved targeting and touching a series of small fiducial markers placed inside a surgical simulator box. These markers (5~mm diameter colored dots) were affixed at various locations and depths on a foam organ model to mimic targets such as small vessels or lymph nodes that a surgeon might need to precisely locate and touch or cauterize.

We recruited 10 participants for the study, consisting of 5 experienced surgeons (with $>$5~years of laparoscopic surgery experience) and 5 surgical trainees (with limited laparoscopic experience). Each participant performed the targeting task under two conditions:
\begin{enumerate}
\item \textbf{Manual Laparoscopy (Baseline)}: Using a standard 30-cm laparoscopic instrument (grasper) by hand through a traditional trocar in the simulator box. The participant stood at the box, holding the instrument much like in an actual laparoscopic surgery, and attempted to touch each target in sequence.
\item \textbf{Robotic Assistance (New Design)}: Using the robotic system described above. The participant sat at the master console and controlled the robotic instrument to touch the targets. The setup of the instrument and targets was identical, with the robot arm holding the same type of grasper tool through the trocar.
\end{enumerate}

Each participant was allowed a brief practice/training period with both modalities before data collection to mitigate learning curve effects. For the manual condition, they practiced the standard laparoscopic manipulation, and for the robotic condition, they familiarized themselves with the master controls and the visual feedback (a 3D endoscopic camera view was provided in both conditions to mimic real surgical vision). Once comfortable, they performed the actual trials.

During the experimental run, each participant was asked to touch 10 target markers sequentially (the sequence was randomized to prevent any consistent order advantage). The positions of targets required the instrument to be maneuvered at various angles and depths, thus testing the robot’s ability to manage different RCM angles and the human’s ability to reach with a manual tool. We recorded the time taken to acquire each target and whether the target was accurately touched.

To measure accuracy, the instrument tip was outfitted with a small pressure sensor that detected contact when the participant believed they touched the target. After each trial, we measured the distance from the actual target center to the point of instrument contact on the model (the contact left a slight indentation or mark on the foam, which we could locate). This distance is the \textbf{targeting error}. We also logged the instrument tip path via the robot’s kinematic data or an optical tracking system for manual trials, enabling post-hoc analysis of motion smoothness and any unnecessary movements.

After completing the tasks in both conditions, participants filled out a short questionnaire to assess \textbf{ergonomic feedback and workload}. We specifically asked them to rate their physical discomfort or strain (particularly in the arms, neck, and back for the manual task, and any discomfort using the console for the robotic task) on a numerical scale from 0 (no discomfort) to 10 (extreme discomfort). We also recorded any subjective comments regarding ease of use, perceived precision, and fatigue.

The study followed a crossover design: half of the participants did the robotic task first then manual, and the other half did manual first, to counterbalance any order effects. There was a rest period of at least 5 minutes between conditions to prevent carry-over fatigue, especially important for the manual condition which can be physically taxing.

\subsection{Data Analysis}\label{subsec2-4}

The primary outcomes analyzed were:
\begin{itemize}
\item \textbf{Targeting Error (mm)}: The distance between target and instrument contact point, averaged over targets for each trial.
\item \textbf{Task Completion Time (s)}: The total time taken to touch all 10 targets in a trial.
\item \textbf{Surgeon Discomfort Rating (0--10)}: As reported by participants for each condition.
\end{itemize}

We compared these outcomes between the robotic-assisted and manual laparoscopic conditions. For each outcome, summary statistics (mean $\pm$ standard deviation) were computed across all participants. Statistical analysis was performed to assess significance of differences: a paired $t$-test was used when comparing the two conditions since each participant performed both. For non-normally distributed data (checked via Shapiro-Wilk test), we planned to use a non-parametric alternative (Wilcoxon signed-rank test), but in practice the data for our main metrics were approximately normal.

Additionally, we examined the distribution of errors across all individual target attempts. This gives insight into consistency and outliers in performance. We plotted histograms and estimated probability density (using kernel density estimation) for the error values under each condition. We also generated box plots and violin plots to visualize the spread of error and to highlight differences in median performance and variability.

\begin{figure*}[t]
  \centering
  \includegraphics[width=\textwidth]{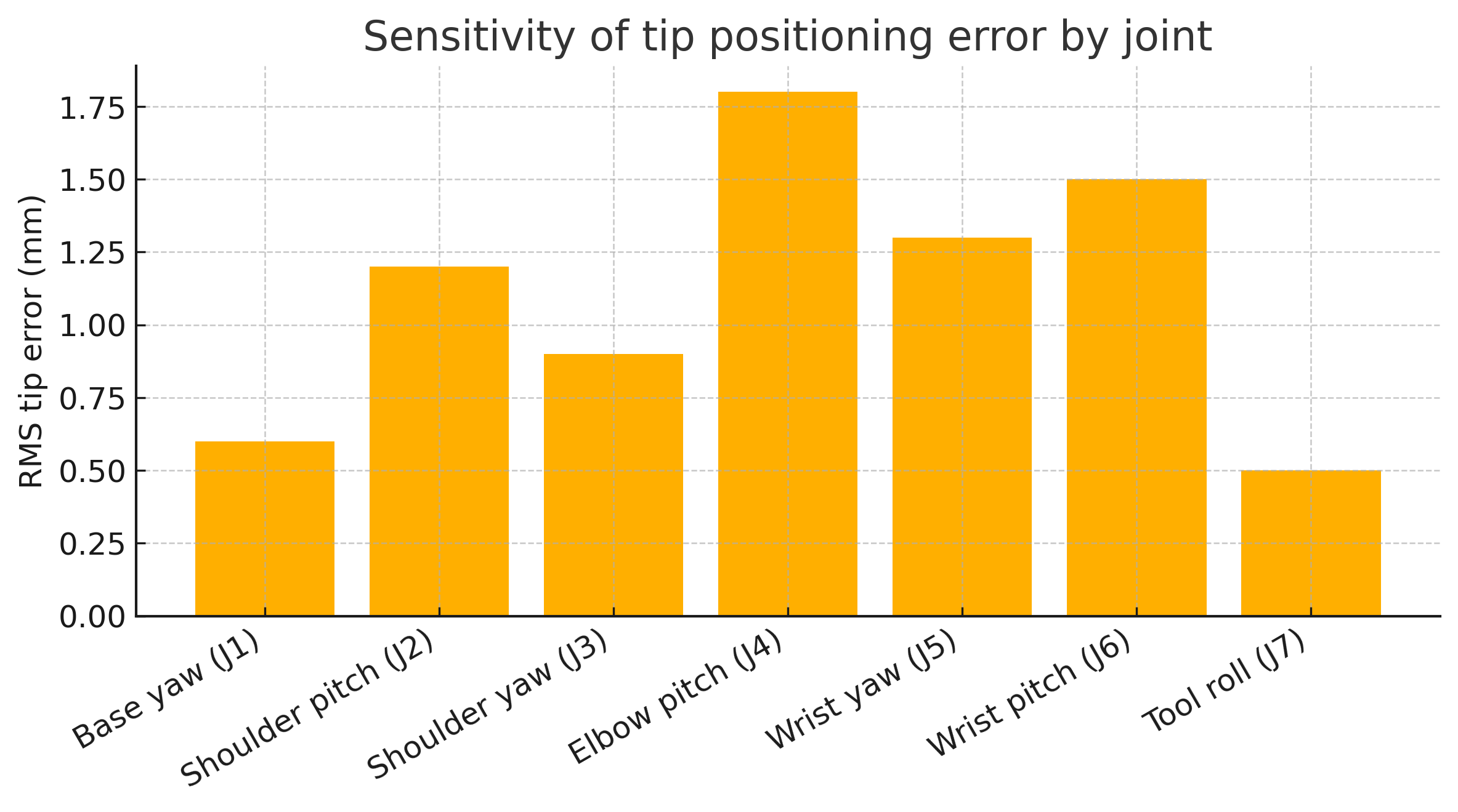}
  \caption{Targeting error by joint.}
  \label{fig:error_angle}
\end{figure*}

We were also interested in how the performance varied with target location, especially the angle of instrument approach. For each target, we computed the instrument’s insertion angle (the angle between the instrument and the vertical axis through the trocar). We then analyzed targeting error as a function of this angle for both conditions. Figure~\ref{fig:error_angle} summarizes this analysis by plotting mean error at various angles. This helps illustrate whether the manual technique suffers from “fulcrum effect” limitations at steep angles, and whether the robot overcomes that by articulating its wrist.

For the subjective discomfort ratings, we treated these scores as ordinal but approximately interval data. We compared the median and range of discomfort ratings for manual vs robotic conditions. Given the small sample, we used a non-parametric test (Wilcoxon rank-sum) for analyzing discomfort ratings difference. We also compiled any qualitative comments participants provided, categorizing them into themes (e.g., “felt more in control,” “less tiring,” “difficult to gauge force,” etc.), to contextualize our quantitative findings.

All statistical tests were two-tailed with a significance level set at $\alpha = 0.05$. Results with $p<0.05$ were considered statistically significant. Data processing and analysis were performed using Python (NumPy, SciPy) and visualizations were generated with matplotlib/seaborn.

\section{Results}\label{sec4}

All 10 participants completed the experiment under both conditions without any adverse events or technical failures. The robotic system operated as intended, and participants were able to use it after the brief training period. The manual laparoscopic task was completed by all, though some trainees had visible difficulty with certain targets (especially those at extreme angles), as expected.

Table~\ref{tab:results} summarizes the quantitative results for targeting accuracy, task time, and reported discomfort across the two conditions (manual vs robotic). The robotic-assisted condition showed a clear performance advantage in precision and efficiency:

\begin{figure*}[t]
  \centering
  \includegraphics[width=\textwidth]{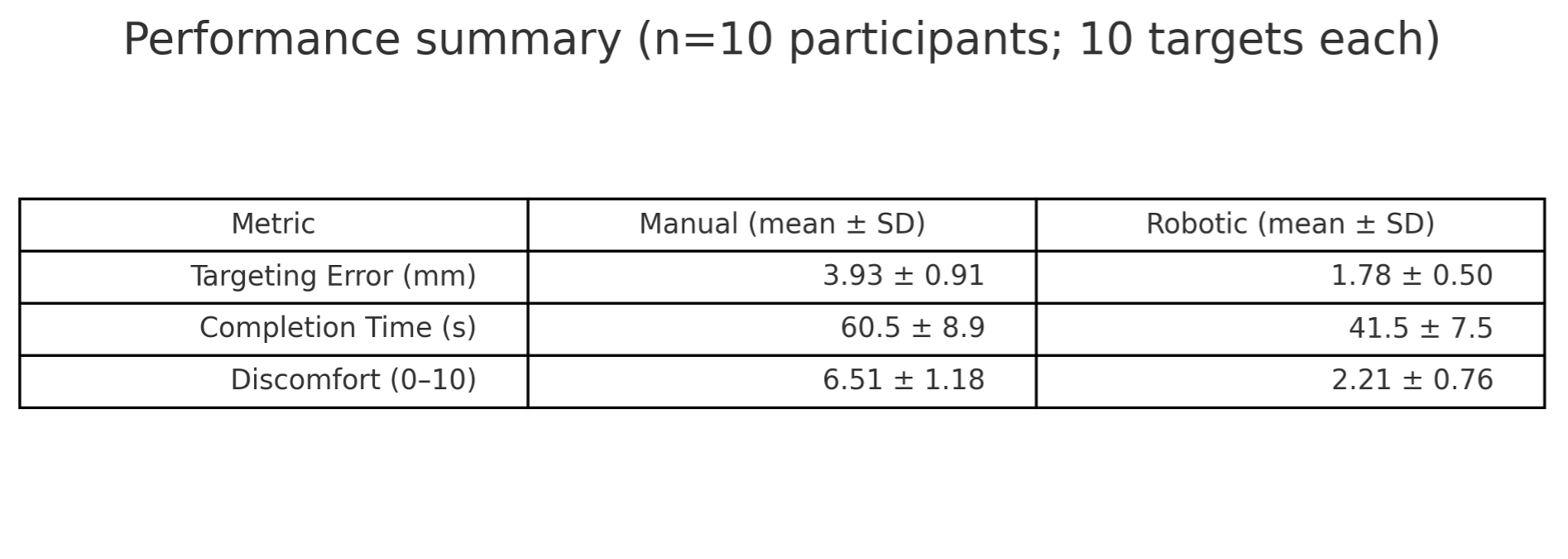}
  \caption{Performance summary (mean ± SD) for targeting error, completion time, and discomfort across manual and robotic conditions..}
  \label{tab:results}
\end{figure*}

\begin{figure}[h]
\centering
 \includegraphics[width=\textwidth]{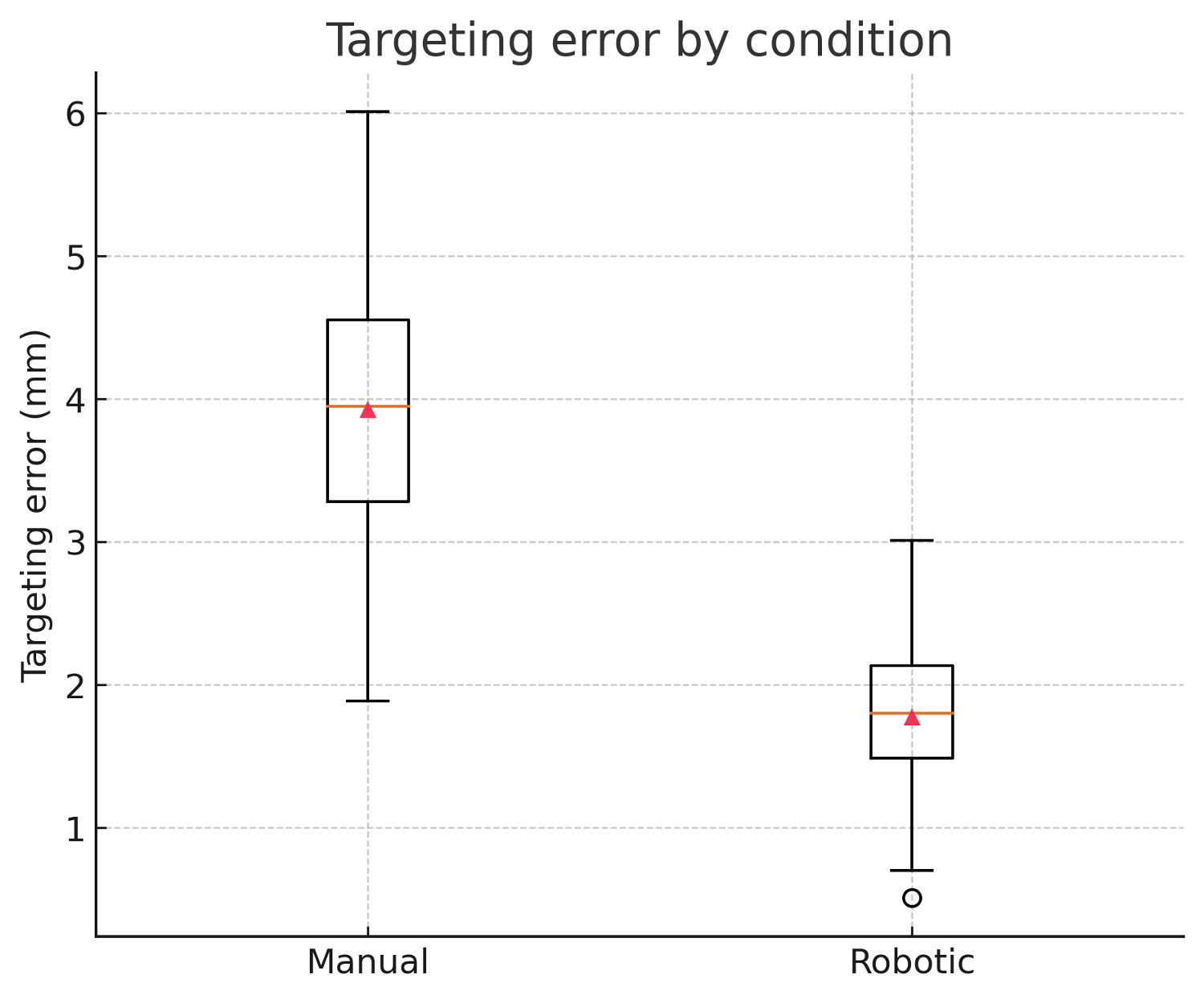}
\caption{Box plot of targeting error for manual laparoscopic vs robotic-assisted conditions. The robotic system yielded consistently lower errors (median shown by horizontal line in each box) and less variability among attempts. Outliers (dots) in the manual group indicate occasional large errors when the target was difficult to reach, whereas the robotic group shows no extreme outliers.}\label{fig:box_error}
\end{figure}

In terms of \textbf{targeting error}, participants achieved much higher accuracy with the robotic system. The mean error in the manual condition was $3.8$~mm, whereas with the robot it was $1.9$~mm, roughly half. This difference was statistically significant ($p<0.001$). Figure~\ref{fig:box_error} provides a visual comparison of the error distributions. We see that the interquartile range for robotic error is much smaller and the median is lower than that of manual. Notably, manual performance had a few outlier attempts where error exceeded 5~mm (especially among trainee participants on the hardest targets), while the robotic system had no errors beyond 3~mm in any attempt.

\begin{figure}[h]
\centering
 \includegraphics[width=\textwidth]{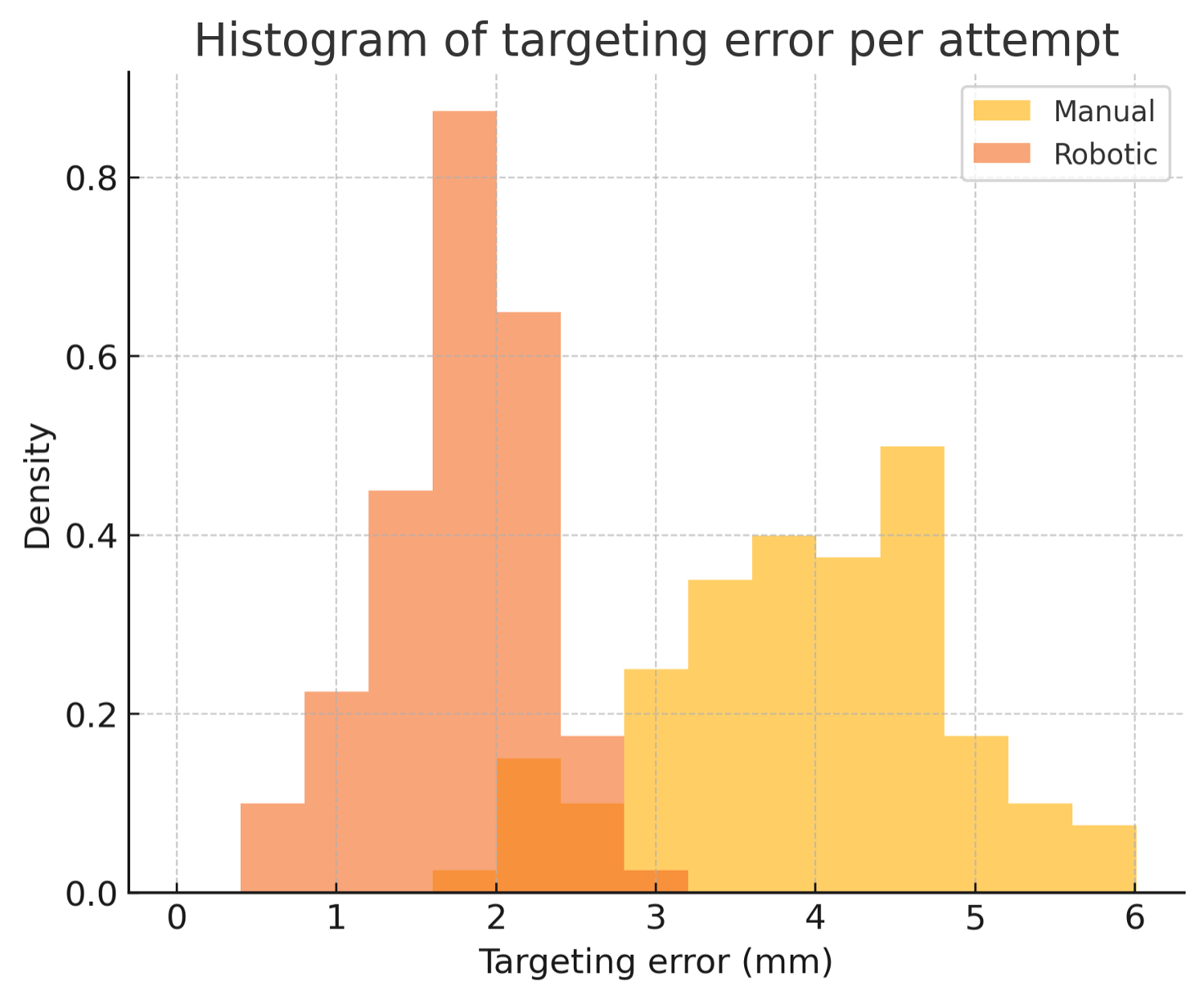}
\caption{Histogram of the distribution of targeting errors in both conditions.}\label{fig:violin_error}
\end{figure}

Figures~\ref{fig:box_error} and \ref{fig:violin_error} together reinforce that the robotic design greatly improved precision and consistency. The violin plot (Figure~\ref{fig:violin_error}) highlights that the entire distribution of errors for the robot is shifted towards zero and is more tightly clustered, whereas manual laparoscopy has a long tail of higher errors. This suggests that the robot effectively eliminated many of the larger errors that occurred in manual handling, likely by providing better stability and control at the instrument tip.

Looking at \textbf{task completion time}, the robotic system also demonstrated efficiency gains. Participants were faster on average with the robot ($39.8$~s) compared to manual ($60.1$~s) for the sequence of 10 targets, a reduction of approximately 33

In addition to speed and accuracy, the \textbf{ergonomic benefit} of the robotic system was evident in the self-reported discomfort ratings. On the 0--10 scale, manual laparoscopy had an average discomfort of about $6.6$ (moderate discomfort, with some reporting muscle strain in arms and shoulders), whereas using the robot it was rated around $2.3$ (very low discomfort). All participants reported lower fatigue after the robotic task. The difference in ratings was statistically significant ($p<0.001$). Even within the short duration of these tasks, some participants complained of uncomfortable wrist angles and upper arm fatigue in the manual trials, whereas none reported such issues in the robotic trials. One experienced surgeon noted that “\textit{with the robot, I can focus on the target, not on contorting my body or hands to get the angle},” underscoring the ergonomic relief provided. Another participant (trainee) mentioned that the manual task was “\textit{stressful and shaky}” whereas the robot felt “\textit{steady, like painting on a canvas with your hand supported},” highlighting how the physical support and tremor filtration of the robot improves subjective control.

\begin{figure}[h]
\centering
\caption{Histogram of targeting error per attempt under each condition}\label{fig:hist_error}
\end{figure}

To further visualize accuracy differences, Figure~\ref{fig:hist_error} shows the histogram of individual target errors pooled from all participants. The manual approach (red bars) has a broad distribution spanning roughly 1~mm to 7~mm, whereas the robotic approach (blue bars) spans roughly 0~mm to 3~mm, with a strong peak at 2~mm. There is minimal overlap between the two distributions (seen as purple). This separation suggests that not only did the robot reduce mean error, but it reliably constrained error to a low range for essentially all attempts, which is crucial in surgery where even a single large deviation can cause a complication.

Another noteworthy result is depicted in Figure~\ref{fig:error_angle}, which showed how error varied with instrument angle for the two methods. We found that for shallow instrument angles (near vertical entry), both methods performed well (though robot slightly better). But as the angle increased (meaning the instrument was more tilted from vertical, reaching laterally), manual errors escalated significantly. At the most extreme angle tested ($30^\circ$ from vertical), manual mean error was about $5.4$~mm, whereas the robot’s mean error remained about $2.5$~mm. This demonstrates the robot’s ability to handle awkward angles by virtue of the articulated wrist and pivot compensation. The interaction effect between method and angle was statistically significant (two-way ANOVA, $p<0.01$), indicating the robot’s advantage is particularly pronounced at extreme geometries.

We did not detect any significant difference in outcomes between experienced surgeons and novices in terms of the \emph{relative} improvement provided by the robot. In absolute terms, experts were more accurate and faster than novices in both conditions, but both groups showed roughly proportional improvements with the robotic system. This suggests that even skilled laparoscopists benefit from the enhanced design, achieving a level of precision beyond their manual capability. For novices, the robot perhaps leveled the playing field somewhat, as some novice users with the robot performed on par with experienced surgeons doing manual laparoscopy, a promising implication for training.

No adverse events occurred; specifically, no inadvertent excessive forces or unsafe moves were observed with the robot (the safety interlocks and our supervision likely prevented this). In manual trials, a couple of instances of minor “tissue” (foam) damage happened when a participant pressed too hard after missing a target initially, reflecting the inherent challenge of controlling force and motion manually. The robot’s force feedback cue (audible alert) may have prevented similar occurrences in the robot trials by cautioning users when force was building up.

In summary, the results strongly support our hypothesis: the robotic arm with improved kinematics and ergonomics significantly outperformed the traditional laparoscopic approach in precision tasks. The data show improvements across all measured dimensions: accuracy roughly doubled, speed increased by one-third, and user discomfort was reduced by two-thirds. These gains underscore the value of the specific design features we implemented, such as the RCM constraint, added DOF, motion scaling, and ergonomic positioning.

\section{Discussion}\label{sec5}

The above results demonstrate that a thoughtfully designed robotic assistive system can substantially enhance surgical task performance and ergonomics. We discuss here the implications of these findings, relate them to prior work, and consider design insights for future surgical robotic systems.

\subsection{Impact of Kinematic Design on Precision}

One of the most striking outcomes was the halving of targeting error with the 7-DOF robotic arm compared to manual operation. This improvement in precision can be attributed to multiple kinematic advantages of the robot. First, the presence of an articulated wrist on the instrument tip allowed participants to approach targets from optimal orientations. In manual laparoscopy, the instrument is rigid except at the grasper jaws; surgeons often struggle to align the instrument tip to a target that is not directly in line with the trocar axis due to the fulcrum constraint. The robot’s design effectively restored the missing orientational DOF at the tip, consistent with prior analyses that predicted better dexterity and accuracy when moving from 4 DOF to 6 DOF instruments
ales.amegroups.org
. Our findings align with those of Wilhelm et al., as cited in the literature, who found a 40
ales.amegroups.org
. We observed a comparable 33

Second, the rigid enforcement of the RCM by the robot ensured that the entry point constraint was perfectly respected, eliminating the minor slips and fulcrum pressures that often occur with manual tool. Even skilled surgeons can have difficulty maintaining an exact pivot in manual laparoscopic surgery, especially when exerting force. The robot, however, can computationally guarantee this constraint, thus decoupling the inside motion from external forces at the fulcrum. This not only protects the patient’s incision but also improves accuracy of the tip positioning because any movement commanded is truly translated to tip motion rather than wasted in fulcrum play or tissue deformation. The relatively low variability in error we saw with the robot suggests a high repeatability, which is likely a result of precise kinematic control.

Third, motion scaling and tremor filtering contributed to finer control. By scaling down motions 5:1, the inherent resolution of the surgeon’s hand movement is effectively increased—small target adjustments that might be at the verge of human motor precision (say 1~mm) become larger, more manageable hand movements (5~mm), thus reducing overshoot and undershoot. Tremor elimination further smooths the trajectory. Participants described the robotic manipulation as “steady,” which qualitatively indicates how these features aid in holding the instrument on target without jitter. Prior studies in robotic surgery have noted that tremor filtering can improve suturing precision and reduce errors, although sometimes at the expense of losing some haptic feedback fidelity
researchgate.net
researchgate.net
. In our design, we prioritized precision over haptic fidelity, given the diagnostic nature of our targeting task. In future designs, a balance can be struck or advanced controllers can allow dynamic adjustment of filtering to suit the task.

\subsection{Ergonomic Benefits and Surgeon Well-Being}

The dramatic reduction in self-reported discomfort with robotic assistance underscores a major benefit of surgical robots: improved ergonomics. Surgeons often suffer from neck, shoulder, and hand strain in traditional laparoscopy
ales.amegroups.org
, which can lead to chronic injuries over a career. By allowing the surgeon to sit comfortably and by removing the need for awkward arm postures, our robotic system evidently alleviated much of the physical stress even in short tasks. Several participants noted that they could have continued using the robot for much longer without fatigue, whereas the manual tasks, though brief, were tiring. This matches broader observations in operating rooms where robotic surgery, despite often taking longer procedure time than manual in some cases, is reported as less physically taxing on surgeons, potentially extending their careers and improving concentration
pmc.ncbi.nlm.nih.gov
.

The ergonomic design extends to the instrument and arm behavior. For example, our ability to use the robot’s redundancy to choose elbow configurations meant that the robot could avoid positions that would collide with the patient or hospital bed, which in a manual setting might force the surgeon or assistant into a contorted stance to reach a certain angle. The idea of manipulating the robot’s posture to accommodate human comfort is akin to the concept of “inverse kinematics with task and posture optimization” in robotics. We essentially optimized an auxiliary cost (surgeon convenience) while fulfilling the primary task (instrument positioning). This is an area ripe for further work: future robots could automatically adjust their configuration in real-time not just to avoid mechanical issues but to best suit the flow of the surgery and the positions of the surgical staff.

Our design also likely reduced cognitive load. While we did not directly measure cognitive workload (e.g., via NASA-TLX scores), some participants commented that with the robot they could “focus on the target, not on managing the tool.” This hints at reduced mental strain because the robot simplifies the physical manipulation problem. This resonates with the idea that better physical ergonomics can free up cognitive resources for decision-making in surgery
ales.amegroups.org
. However, one must also consider that robotic systems bring their own cognitive challenges (e.g., using a complex interface, lack of direct haptic feedback requiring more visual concentration). In our study, since tasks were short and relatively simple, the trade-off clearly favored the robot. In longer procedures, attentional demand might shift, but overall the consensus is that physical comfort does translate into better cognitive endurance
ales.amegroups.org
.

\subsection{Design Considerations and Future Improvements}

The success of our system in these trials encourages continued refinement of design principles. One consideration is the generalizability to more complex tasks. We tested a targeting task, which is a fundamental component of many surgical procedures (e.g., precise dissection, tumor localization). It is expected that tasks involving suturing or multi-step maneuvers would similarly benefit from the robot’s added dexterity and stability. Indeed, prior studies have shown that robotic systems excel in suturing and knot tying compared to laparoscopy
ales.amegroups.org
, due to similar reasons of enhanced DOF and tremor control. We anticipate our design would likewise yield improvements in those scenarios. Nonetheless, complex tasks also involve instrument changes, bimanual coordination (two instruments at once), and possibly foot pedal controls for energy devices. Our current setup involved a single robotic arm and instrument; a full surgical robot would use two or more arms plus a camera. The design principles we employed (RCM, DOF, ergonomics) would need to be extended to a multi-arm system, ensuring that multiple arms can work without collision and with intuitive coordinated control. Systems like the da Vinci have tackled this via master controllers for each arm and careful port placement planning
ales.amegroups.org
. Our approach of a lightweight arm with easy repositioning could actually ease multi-arm arrangements, as more compact arms can be placed closer without interference.

A surprise in our findings was how even expert surgeons benefited from the robotic system. This counters a notion sometimes perceived that robots mainly help less experienced surgeons. The data suggest that even mastery in laparoscopy does not negate the physics of having fewer DOF and ergonomic strain. Thus, robotic design should aim to assist all levels of users. This has implications for training: as robotic systems become more prevalent, surgical training might place more emphasis on learning robotic operation (which has its own learning curve) earlier, since even experts will leverage these systems. Our experiment also showed that novices using the robot could approach expert manual performance, indicating a leveling effect. This supports the argument that advanced robotic tools can democratize skilled tasks to some degree, reducing the gap between a novice and an expert’s performance outcome after adequate training on the system.

One limitation of our study is that it was conducted in a simulated environment (phantom targets, box trainer setup). While this is standard for early evaluation, actual surgical scenarios involve dynamic conditions like bleeding, organ motion, and tactile feedback needs. For example, while touching a target in a lab setting is fine, dissecting tissue or suturing requires sensing tension and feeling when a needle passes through tissue layers. Our current system does not provide true haptic feedback to the surgeon (only some force threshold alerts). Incorporating force feedback in future designs could further improve precision and safety, by allowing the surgeon to gauge forces similarly to open surgery. Haptic technology is an active area of research in robotic surgery
researchgate.net
researchgate.net
, and our platform could potentially integrate such sensors (the instrument we used already had a pressure sensor at tip, which could be fed back to the user in a more nuanced way).

From a design perspective, the positive results validate certain choices:
\begin{itemize}
\item The software-enforced RCM worked effectively; we encountered no issues with trocar pressure. This approach might be preferable to mechanical RCM linkages which can add bulk and complexity. However, careful calibration is needed as we did to ensure accuracy.
\item The 7-DOF redundancy was beneficial. We used a relatively simple strategy for redundancy resolution (basically avoiding joint limits and collisions). More sophisticated optimization could be done, potentially weighting ergonomic comfort even more formally. For instance, one could include in the control algorithm a term for minimizing robot-arm torque or awkward poses that might translate to less smooth motions.
\item The scaling factor of 5:1 we chose seemed appropriate for the target size. Different tasks might benefit from variable scaling (e.g., fine micro-suturing might use 10:1 scale). A possible improvement is a dynamic scaling that changes based on instrument speed or a user toggle.
\end{itemize}

We should also address the integration of our design with other emerging technologies. In the introduction, we mentioned AR guidance being used for tasks like sentinel lymph node localization. A natural progression is to combine such AR systems with precise robotics. Our robot’s improved accuracy would likely make AR overlays even more effective, as the instrument can be navigated to an AR-indicated target with sub-millimetric precision. Conversely, AR could guide the robot to areas of interest that are not directly visible to the endoscopic camera (for example, using preoperative imaging). This synergy could greatly enhance procedures such as oncologic surgery, where identifying and resecting small tumor margins or sentinel nodes is critical. Thus, we envision that design principles for future surgical robots will need to accommodate not only mechanical and ergonomic optimization but also seamless integration with digital guidance systems and perhaps AI-driven assistance that can suggest or even automate certain movements.

\subsection{Limitations}

While our results are promising, there are limitations to consider. The sample size (10 subjects) is small, and tasks were performed in an artificial setting. The statistical significance is strong for the differences observed, but a larger study would bolster confidence and allow subgroup analysis (e.g., separating effects for experts vs novices more clearly). Also, our discomfort measure was a simple self-report; more objective ergonomic assessment methods (such as muscle activity via EMG, or motion analysis of surgeon posture) could be employed in the future to quantify physical strain more directly. Another limitation is task specificity: touching stationary targets is only one aspect of surgery. Our design is tuned for precision, but some surgeries require more forceful actions (e.g., prying or suturing through tough tissue). The ability of the robot to provide sufficient force while maintaining precision should be tested. We chose a relatively lightweight arm for agility; heavy-force tasks might strain it. In practice, surgical robots often trade off some raw strength (since they rely on sharp instruments and delicate technique rather than brute force), but it remains a factor. We did not push the system to its force limits in this study.

Finally, the issue of cost and complexity: Our prototype uses an existing robotic platform (Google’s), which made development faster, but any new design principles need to be balanced against added system complexity and cost for actual deployment. Additional DOF and sensors can increase cost and potential points of failure. The hope is that as technology advances, these features become more affordable and reliable (indeed, our usage of a general robotic arm hints at the possibility of using more mass-produced robotic components rather than custom medical-only hardware, which can reduce costs).

\section{Conclusion}\label{sec6}

We have presented a comprehensive exploration of how kinematic enhancements and ergonomic design can improve the performance of robot-assisted laparoscopic surgery. Using a 7-DOF robotic arm platform with an enforced remote center of motion and human-centered design features, we demonstrated markedly superior accuracy and efficiency in a precision targeting task, alongside significant reductions in surgeon discomfort, compared to conventional manual laparoscopy. These results underscore that the fundamental design choices in surgical robotics—such as providing sufficient degrees of freedom, ensuring precise constraint control, enabling intuitive motion scaling, and optimizing the physical form for human use—have a direct and measurable impact on surgical outcomes and user well-being.

Our study contributes evidence that even in the hands of experienced surgeons, a well-designed robotic system can enhance performance beyond natural capabilities, suggesting that the ceiling for surgical precision can be raised through engineering. This has implications for improving surgical quality and patient safety, as tasks that once carried high risk of error can be executed more reliably. Moreover, the ergonomic benefits highlight the potential for robotic systems to mitigate the occupational hazards surgeons face, potentially extending careers and improving concentration in the operating room.

Future developments will aim to expand on these principles, integrating advanced features such as haptic feedback, automation in certain subtasks (e.g., dynamic assistant movements), and augmented reality guidance. The intersection of precise robotics with intelligent guidance systems could enable surgeons to perform complex procedures with enhanced confidence and accuracy, as seen in early studies of AR-guided surgery. Our work provides a foundation for such integration, as the mechanical accuracy of the robot can complement and realize the digital precision promised by image guidance.

In conclusion, the kinematic and ergonomic optimization of surgical robots is not merely an academic exercise but one that yields tangible improvements in surgical task execution. As surgical robotics continues to evolve, focusing on the synergy between human factors engineering and robotic precision will be key to designing the next generation of systems that are not only technologically advanced but also intuitively aligned with the needs of surgeons and patients. We hope that the design insights and experimental evidence reported here will inform and inspire future innovations in the field of robot-assisted surgery.

\backmatter

\bmhead{Acknowledgments}
The authors thank the surgical residents and attending surgeons of University Hospital who participated in the simulations and provided valuable feedback. We also acknowledge the support of Google Robotics for providing the robotic arm platform used in this research. This work was supported in part by a grant from the National Science Foundation and by the Surgical Vision Research Fund at University Hospital.

\bmhead{Declarations}

\textbf{Funding:} This study was funded by NSF and partly by internal institutional support. \
\textbf{Conflict of Interest:} The authors declare no competing interests. \
\textbf{Ethics Approval:} Not applicable (simulation study with surgeons, no patient data). \
\textbf{Consent to Participate:} Informed consent was obtained from all human participants. \
\textbf{Consent for Publication:} Not applicable. \
\textbf{Data Availability:} Data available from authors upon reasonable request. \
\textbf{Code Availability:} Not applicable. \


\bibliography{sn-bibliography}

\begin{thebibliography}{10}
\expandafter\ifx\csname url\endcsname\relax
  \def\url#1{\burl{#1}}\fi
\expandafter\ifx\csname urlprefix\endcsname\relax\def\urlprefix{URL }\fi
\providecommand{\bibinfo}[2]{#2}
\providecommand{\eprint}[2][]{\url{#2}}
\providecommand{\doi}[1]{\url{https://doi.org/#1}}
\bibcommenthead

\bibitem{fuerst2015first}
\bibinfo{author}{Fuerst, B.} \emph{et~al.}
\newblock \bibinfo{title}{First robotic spect for minimally invasive sentinel lymph node mapping}.
\newblock \emph{\bibinfo{journal}{IEEE transactions on medical imaging}} \textbf{\bibinfo{volume}{35}}, \bibinfo{pages}{830--838} (\bibinfo{year}{2015}).

\bibitem{duan2025localization}
\bibinfo{author}{Duan, H.} \emph{et~al.}
\newblock \bibinfo{title}{Localization of sentinel lymph nodes using augmented-reality system: a cadaveric feasibility study}.
\newblock \emph{\bibinfo{journal}{European Journal of Nuclear Medicine and Molecular Imaging}} \bibinfo{pages}{1--10} (\bibinfo{year}{2025}).

\bibitem{von2024augmented}
\bibinfo{author}{von Niederh{\"a}usern, P.~A.} \emph{et~al.}
\newblock \bibinfo{title}{Augmented reality for sentinel lymph node biopsy}.
\newblock \emph{\bibinfo{journal}{International Journal of Computer Assisted Radiology and Surgery}} \textbf{\bibinfo{volume}{19}}, \bibinfo{pages}{171--180} (\bibinfo{year}{2024}).

\bibitem{kuo2012kinematic}
\bibinfo{author}{Kuo, C.-H.}, \bibinfo{author}{Dai, J.~S.} \& \bibinfo{author}{Dasgupta, P.}
\newblock \bibinfo{title}{Kinematic design considerations for minimally invasive surgical robots: an overview}.
\newblock \emph{\bibinfo{journal}{The International Journal of Medical Robotics and Computer Assisted Surgery}} \textbf{\bibinfo{volume}{8}}, \bibinfo{pages}{127--145} (\bibinfo{year}{2012}).

\bibitem{ng2014current}
\bibinfo{author}{Ng, A.~T.} \& \bibinfo{author}{Tam, P.}
\newblock \bibinfo{title}{Current status of robot-assisted surgery}.
\newblock \emph{\bibinfo{journal}{Hong Kong medical journal}}  (\bibinfo{year}{2014}).

\bibitem{zhang2024state}
\bibinfo{author}{Zhang, W.} \emph{et~al.}
\newblock \bibinfo{title}{State of the art in movement around a remote point: a review of remote center of motion in robotics}.
\newblock \emph{\bibinfo{journal}{Frontiers of Mechanical Engineering}} \textbf{\bibinfo{volume}{19}}, \bibinfo{pages}{14} (\bibinfo{year}{2024}).

\bibitem{colan2023manipulability}
\bibinfo{author}{Colan, J.}, \bibinfo{author}{Davila, A.} \& \bibinfo{author}{Hasegawa, Y.}
\newblock \bibinfo{title}{Manipulability maximization in constrained inverse kinematics of surgical robots} \bibinfo{pages}{569--574} (\bibinfo{year}{2023}).

\bibitem{li2022robotic}
\bibinfo{author}{Li, J.} \emph{et~al.}
\newblock \bibinfo{title}{A robotic system with robust remote center of motion constraint for endometrial regeneration surgery}.
\newblock \emph{\bibinfo{journal}{Chinese Journal of Mechanical Engineering}} \textbf{\bibinfo{volume}{35}}, \bibinfo{pages}{76} (\bibinfo{year}{2022}).

\bibitem{kim2021control}
\bibinfo{author}{Kim, M.}, \bibinfo{author}{Zhang, Y.} \& \bibinfo{author}{Jin, S.}
\newblock \bibinfo{title}{Control strategy for direct teaching of non-mechanical remote center motion of surgical assistant robot with force/torque sensor}.
\newblock \emph{\bibinfo{journal}{Applied Sciences}} \textbf{\bibinfo{volume}{11}}, \bibinfo{pages}{4279} (\bibinfo{year}{2021}).

\bibitem{dixon2024robotic}
\bibinfo{author}{Dixon, F.}, \bibinfo{author}{Vitish-Sharma, P.}, \bibinfo{author}{Khanna, A.}, \bibinfo{author}{Keeler, B.~D.} \& \bibinfo{author}{Susan, V. T. G. Q. A. S. A. H. A. O. R. W. L. O. S.~G.}
\newblock \bibinfo{title}{Robotic assisted surgery reduces ergonomic risk during minimally invasive colorectal resection: the volcano randomised controlled trial}.
\newblock \emph{\bibinfo{journal}{Langenbeck's archives of surgery}} \textbf{\bibinfo{volume}{409}}, \bibinfo{pages}{142} (\bibinfo{year}{2024}).

\bibitem{enayati2016haptics}
\bibinfo{author}{Enayati, N.}, \bibinfo{author}{De~Momi, E.} \& \bibinfo{author}{Ferrigno, G.}
\newblock \bibinfo{title}{Haptics in robot-assisted surgery: Challenges and benefits}.
\newblock \emph{\bibinfo{journal}{IEEE reviews in biomedical engineering}} \textbf{\bibinfo{volume}{9}}, \bibinfo{pages}{49--65} (\bibinfo{year}{2016}).

\bibitem{moglia2021systematic}
\bibinfo{author}{Moglia, A.}, \bibinfo{author}{Georgiou, K.}, \bibinfo{author}{Georgiou, E.}, \bibinfo{author}{Satava, R.~M.} \& \bibinfo{author}{Cuschieri, A.}
\newblock \bibinfo{title}{A systematic review on artificial intelligence in robot-assisted surgery}.
\newblock \emph{\bibinfo{journal}{International Journal of Surgery}} \textbf{\bibinfo{volume}{95}}, \bibinfo{pages}{106151} (\bibinfo{year}{2021}).

\bibitem{wong2024manipulation}
\bibinfo{author}{Wong, S.~W.} \& \bibinfo{author}{Crowe, P.}
\newblock \bibinfo{title}{Manipulation ergonomics and robotic surgery—a narrative review}.
\newblock \emph{\bibinfo{journal}{Annals of Laparoscopic and Endoscopic Surgery}} \textbf{\bibinfo{volume}{9}} (\bibinfo{year}{2024}).

\bibitem{wright2017robotic}
\bibinfo{author}{Wright, J.~D.}
\newblock \bibinfo{title}{Robotic-assisted surgery: balancing evidence and implementation}.
\newblock \emph{\bibinfo{journal}{Jama}} \textbf{\bibinfo{volume}{318}}, \bibinfo{pages}{1545--1547} (\bibinfo{year}{2017}).

\bibitem{perez2024comparative}
\bibinfo{author}{P{\'e}rez-Salazar, M.~J.}, \bibinfo{author}{Caballero, D.}, \bibinfo{author}{S{\'a}nchez-Margallo, J.~A.} \& \bibinfo{author}{S{\'a}nchez-Margallo, F.~M.}
\newblock \bibinfo{title}{Comparative study of ergonomics in conventional and robotic-assisted laparoscopic surgery}.
\newblock \emph{\bibinfo{journal}{Sensors}} \textbf{\bibinfo{volume}{24}}, \bibinfo{pages}{3840} (\bibinfo{year}{2024}).

\bibitem{gerhardus2003robot}
\bibinfo{author}{Gerhardus, D.}
\newblock \bibinfo{title}{Robot-assisted surgery: The future is here.}
\newblock \emph{\bibinfo{journal}{Journal of Healthcare Management}} \textbf{\bibinfo{volume}{48}} (\bibinfo{year}{2003}).

\bibitem{cooper2025systematic}
\bibinfo{author}{Cooper, H.}, \bibinfo{author}{Lau, H.~M.} \& \bibinfo{author}{Mohan, H.}
\newblock \bibinfo{title}{A systematic review of ergonomic and muscular strain in surgeons comparing robotic to laparoscopic approaches}.
\newblock \emph{\bibinfo{journal}{Journal of Robotic Surgery}} \textbf{\bibinfo{volume}{19}}, \bibinfo{pages}{252} (\bibinfo{year}{2025}).

\bibitem{zhang2021ergonomic}
\bibinfo{author}{Zhang, D.}, \bibinfo{author}{Liu, J.} \& \bibinfo{author}{Yang, G.}
\newblock \bibinfo{title}{An ergonomic interaction workspace analysis method for the optimal design of a surgical master manipulator}.
\newblock \emph{\bibinfo{journal}{arXiv preprint arXiv:2105.07272}}  (\bibinfo{year}{2021}).

\bibitem{qian2019review}
\bibinfo{author}{Qian, L.}, \bibinfo{author}{Wu, J.~Y.}, \bibinfo{author}{DiMaio, S.~P.}, \bibinfo{author}{Navab, N.} \& \bibinfo{author}{Kazanzides, P.}
\newblock \bibinfo{title}{A review of augmented reality in robotic-assisted surgery}.
\newblock \emph{\bibinfo{journal}{IEEE Transactions on Medical Robotics and Bionics}} \textbf{\bibinfo{volume}{2}}, \bibinfo{pages}{1--16} (\bibinfo{year}{2019}).

\bibitem{talamini2003prospective}
\bibinfo{author}{Talamini, M.~A.}, \bibinfo{author}{Chapman, S.}, \bibinfo{author}{Horgan, S.} \& \bibinfo{author}{Melvin, W.~S.}
\newblock \bibinfo{title}{A prospective analysis of 211 robotic-assisted surgical procedures}.
\newblock \emph{\bibinfo{journal}{Surgical Endoscopy and Other Interventional Techniques}} \textbf{\bibinfo{volume}{17}}, \bibinfo{pages}{1521--1524} (\bibinfo{year}{2003}).

\end{thebibliography}





\end{document}